\documentclass{article}

\usepackage[nonatbib, final]{neurips}

\usepackage[utf8]{inputenc} 
\usepackage[T1]{fontenc}    
\usepackage{hyperref}       
\usepackage{url}            
\usepackage{booktabs}       
\usepackage{amsfonts}       
\usepackage{amsmath}
\usepackage{amsthm}
\usepackage{bbm}
\usepackage{multirow}
\usepackage{placeins}
\usepackage[pdftex]{graphicx}

\newcommand{\beginsupplement}{%
        \setcounter{table}{0}
        \renewcommand{\thetable}{S\arabic{table}}%
        \setcounter{figure}{0}
        \renewcommand{\thefigure}{S\arabic{figure}}%
     }

\newtheorem{theorem}{Theorem}

\usepackage{stackengine}

\title{Deep Neural Networks for Rank-Consistent Ordinal Regression Based On Conditional Probabilities}

\author{%
	Xintong Shi\textsuperscript{1} \quad Wenzhi Cao\textsuperscript{1} \quad Sebastian Raschka\textsuperscript{1,2}\thanks{Corresponding author: mail@sebastianraschka.com}\\
	$^{1}$Department of Statistics, University of Wisconsin-Madison \quad ${^2}$Lightning AI\\
}

\begin{document}

\maketitle

\begin{abstract}
In recent times, deep neural networks achieved outstanding predictive performance on various classification and pattern recognition tasks. However, many real-world prediction problems have ordinal response variables, and this ordering information is ignored by conventional classification losses such as the multi-category cross-entropy.  Ordinal regression methods for deep neural networks address this. One such method is the CORAL method, which is based on an earlier binary label extension framework and achieves rank consistency among its output layer tasks by imposing a weight-sharing constraint. However, while earlier experiments showed that CORAL's rank consistency is beneficial for performance, {it is limited by a weight-sharing constraint in a neural network's fully connected output layer, which may restrict the expressiveness and capacity of a network trained using CORAL. We propose a new method for rank-consistent ordinal regression without this limitation. Our rank-consistent ordinal regression framework (CORN) achieves rank consistency by a novel training scheme. This training scheme uses} conditional training sets to obtain the unconditional rank probabilities through applying the chain rule for conditional probability distributions. Experiments on various datasets demonstrate the efficacy of the proposed method to utilize the ordinal target information, and the absence of the weight-sharing restriction improves the performance substantially compared to the CORAL reference approach. Additionally, the suggested CORN method is not tied to any specific architecture and can be utilized with any deep neural network classifier to train it for ordinal regression tasks.
\end{abstract}

\section{Introduction}
\label{sec:introduction}

Many real-world prediction tasks involve ordinal target labels. Popular examples of such ordinal tasks are customer ratings (e.g., a product rating system from 1 to 5 stars) and medical diagnoses (e.g., disease severity labels such as \textit{none}, \textit{mild}, \textit{moderate}, and \textit{severe}). While we can apply conventional classification losses, such as the multi-category cross-entropy, to such problems, they are suboptimal since they ignore the intrinsic order among the ordinal targets. For example, for a patient with \textit{severe} disease status, predicting \textit{none} and \textit{moderate} would incur the same loss even though the difference between \textit{none} and \textit{severe} is more significant than the difference between \textit{moderate} and \textit{severe}. Moreover, unlike in metric regression, we cannot quantify the distance between the ordinal ranks. For instance, the difference between a disease status of \textit{none} and \textit{mild} cannot be quantitatively compared to the difference between \textit{mild} and \textit{moderate}. Hence, ordinal regression (also called ordinal classification or ranking learning) can be considered as an intermediate problem between classification and regression. 

\begin{figure*}
\begin{center}
\centerline{\includegraphics[width=0.78\linewidth]{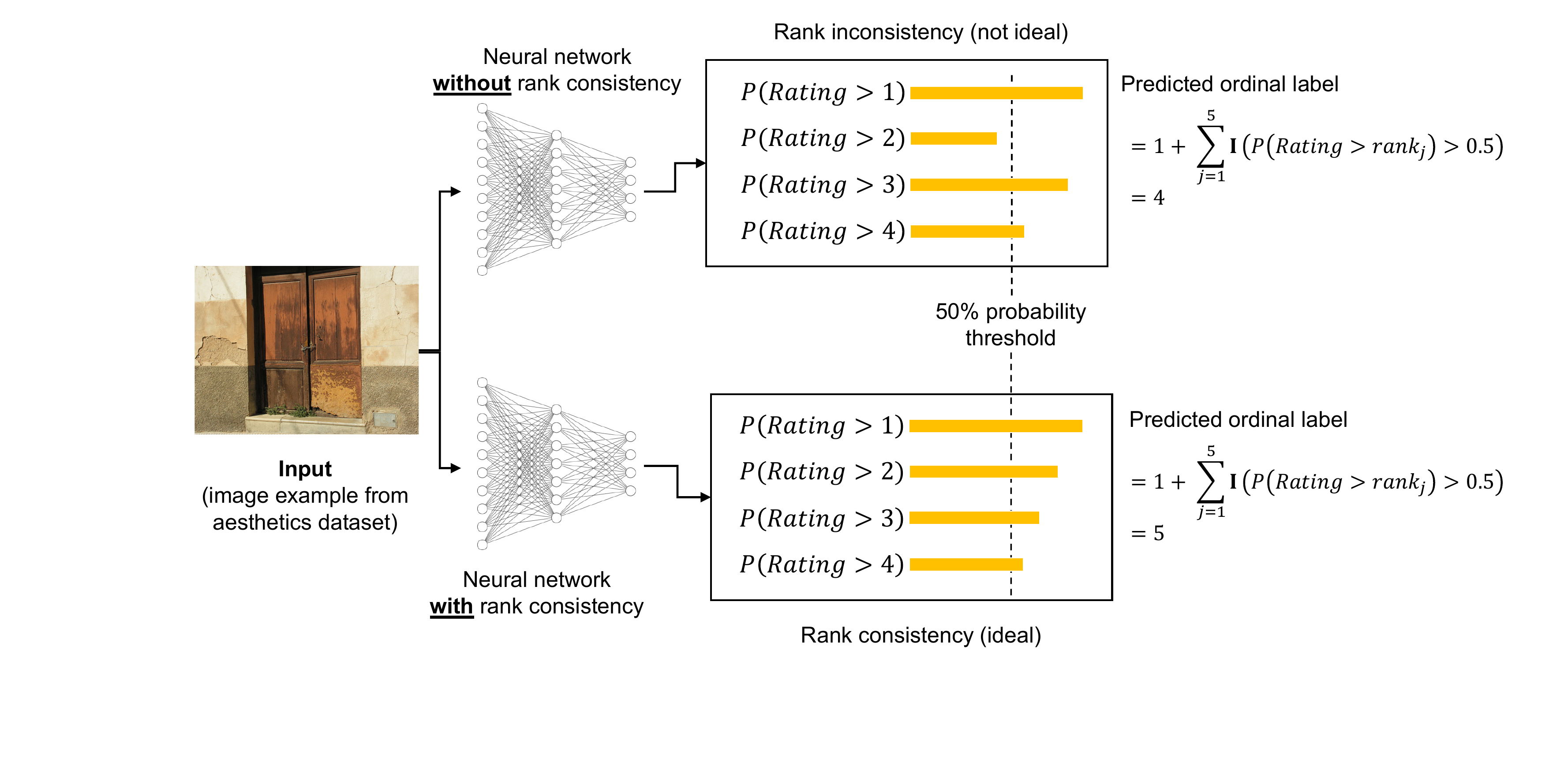}}
\caption{Illustration of the difference between rank-consistent and rank-inconsistent methods.}
\label{fig:overview}
\end{center}
\end{figure*}

Among the most common machine learning-based approaches to ordinal regression is Li and Lin's extended binary classification framework~\cite{li2007ordinal} that was adopted for deep neural networks by Niu et al. in 2016~\cite{niu2016ordinal}. In this work, we solve the rank inconsistency problem (Fig.~\ref{fig:overview}) of this ordinal regression framework without imposing constraints that could limit the expressiveness of the neural network and without substantially increasing the computational complexity.

The contributions of our paper are as follows:
\begin{enumerate}
\item  A new rank-consistent ordinal regression framework, CORN (\textbf{C}onditional \textbf{O}rdinal \textbf{R}egression for \textbf{N}eural Networks), based on the chain rule for conditional probability distributions;
\item  Rank consistency guarantees without imposing the weight-sharing constraint used in the CORAL reference framework~\cite{cao2020rank};
\item  Experiments with different neural network architectures and datasets showing that CORN's removal of the weight-sharing constraint improves the predictive performance compared to the more restrictive reference framework.
\end{enumerate}

\section{Related Work}
\label{sec:related-work}

\subsection{Ordinal Regression Based on Extended Binary Classification Subtasks}
Ordinal regression is a classic problem in statistics, going back to early proportional hazards and proportional odds models~\cite{mccullagh1980regression}. To take advantage of well-studied and well-tuned binary classifiers, the machine learning field developed ordinal regression methods based on extending the rank prediction to multiple binary label classification subtasks~\cite{li2007ordinal}. This approach relies on three steps: (1) extending rank labels to binary vectors, (2) training binary classifiers on the extended labels, and (3) computing the predicted rank label from the binary classifiers. Modified versions of this approach have been proposed in connection with perceptrons~\cite{crammer2002pranking} and support vector machines~\cite{shashua2003ranking,rajaram2003classification,chu2005new}. In 2007, Li and Lin presented a reduction framework unifying these extended binary classification approaches~\cite{li2007ordinal}.

\subsection{Addressing Rank Consistency in Neural Networks for Ordinal Regression}

In 2016, Niu et al. adapted Li and Lin's extended binary classification framework to train deep neural networks for ordinal regression~\cite{niu2016ordinal}; we refer to this method as OR-NN. Across different image datasets, OR-NN was able to outperform other reference methods. However, Niu et al. pointed out that ORD-NN suffers from rank inconsistencies among the binary tasks and that addressing this limitation might raise the training complexity substantially. Cao et al.~\cite{cao2020rank} recently addressed this rank inconsistency limitation via the CORAL method. To avoid increasing the training complexity, CORAL achieves rank consistency by imposing a weight-sharing constraint in the last layer, such that the binary classifiers only differ in their bias units. However, while CORAL outperformed the OR-NN method across several face image datasets for age prediction, the weight-sharing constraint may impose a severe limitation in terms of functions that the neural network can approximate. In this paper, we investigate an alternative approach to guarantee rank consistency without increasing the training complexity and restricting the neural network's expressiveness and capacity.

\subsection{Other Neural Network-Based Methods for Ordinal Regression}

Several deep neural networks for ordinal regression do not build on the extended binary classification framework.  These methods include Zhu et al.'s ~\cite{zhu2021convolutional} convolutional ordinal regression forest for image data, which combines a convolutional neural network with differentiable decision trees. Diaz and Marathe~\cite{diaz2019soft} proposed a soft ordinal label representation obtained from a softmax layer, which can be used for scenarios where interclass distances are known. Another method that does not rely on the extended binary classification framework is Suarez et al.'s distance metric learning algorithm~\cite{suarez2021ordinal}. Petersen et al.~\cite{Petersen2021-diffsort} developed a method based on differentiable sorting networks based on pairwise swapping operations with relaxed sorting operations, which can be used for ranking where the relative ordering is known but the absolute target values are unknown. {Liu et al. adapted pairwise ranking constraints from RankingSVM~\cite{joachims2002optimizing} to reformlate the multi-category loss as a constrained optimization problem for ordinal regression~\cite{liu2018constrained}.}

This paper focuses on addressing the rank inconsistency on OR-NN without imposing the weight-sharing of CORAL~\cite{cao2020rank}, which is why an {exhaustive} study of the methods mentioned above is outside the scope of this paper. However, additional experiments and comparisons with SORD ~\cite{diaz2019soft} and CNNPOR~\cite{liu2018constrained} are included in the Supplementary Material in section \textit{Comparison with Other Deep Learning Methods for Ordinal Regression}.

\section{Proposed Method}
\label{sec:proposed}

This section describes the details of our CORN method, which addresses the rank inconsistency in Niu et al.'s OR-NN~\cite{niu2016ordinal} without requiring CORAL's~\cite{cao2020rank} weight-sharing constraint.

\subsection{Preliminaries}

Let ${D=\left\{\mathbf{x}^{[i]},y^{[i]}\right\}_{i=1}^N}$ denote a dataset for supervised learning consisting of $N$ training examples, where $\mathbf{x}^{[i]}\in \mathcal{X}$ denotes the inputs of the $i$-th training example and $y^{[i]}$ its corresponding class label. In an ordinal regression context, we refer to $y^{[i]}$ as the rank, where ${y^{[i]}\in \mathcal{Y}=\{r_1,r_2,...r_K\}}$ with rank order ${r_K\succ r_{K-1}\succ \ldots\succ r_1}$. The objective of an ordinal regression model is then to find a mapping $h: \mathcal{X}\rightarrow \mathcal{Y}$ that minimizes a loss function $L(h)$. 

\subsection{Motivation}
With CORAL, Cao et al.~\cite{cao2020rank} proposed a deep neural network for ordinal regression that addressed the rank inconsistency of Niu et al.'s OR-NN~\cite{niu2016ordinal}, and experiments showed that addressing rank consistency had a positive effect on predictive performance.

Both CORAL and OR-NN built on an extended binary classification framework~\cite{li2007ordinal}, where  the rank labels are recast into a set of binary tasks, such that ${y^{[i]}_{k} \in \{0,1\}}$ indicates whether $y^{[i]}$ exceeds rank $r_k$. The label predictions are then obtained via $h\left(\mathbf{x}^{[i]}\right)=r_q$, where $q \in \{1, 2, ..., K\}$ is the rank index, which is computed as
\begin{equation} 
\label{eq:rank}
q =  1 + \sum_{k=1}^{K-1} \mathbbm{1}\left\{f_k\left(\mathbf{x}^{[i]}\right) > 0.5\right\}.
\end{equation}

\noindent Here, $f_k\left(\mathbf{x}^{[i]}\right)\in [0,1]$ is the probability prediction of the $k$-th binary classifier in the output layer, and $\mathbbm{1}\{\cdot\}$ is an indicator function that returns $1$ if the inner condition is true and $0$ otherwise.

The CORAL method ensures that the $\{f_k\}_{k=1}^{K-1}$ predictions are rank-monotonic, that is, ${f_1\left(\mathbf{x}^{[i]}\right) \ge f_2\left(\mathbf{x}^{[i]}\right) \ge \dots \ge f_{K-1}\left(\mathbf{x}^{[i]}\right)}$, which provides rank consistency to the ordinal regression model.  While the rank label calculation via Eq.~\ref{eq:rank} does not strictly require consistency among the $K-1$ task predictions, $f_k\left(\mathbf{x}^{[i]}\right)$, it is intuitive to see why rank consistency can be theoretically beneficial and can lead to more interpretable results via the binary subtasks. While CORAL provides this rank consistency, CORAL's limitation is a weight-sharing constraint in the output layer. Consequently, all binary classification tasks use the same weight parameters and only differ in their bias units, which may limit the flexibility and expressiveness of an ordinal regression neural network based on CORAL.

\begin{figure*}[htb!]
\begin{center}
\centerline{\includegraphics[width=0.98\linewidth]{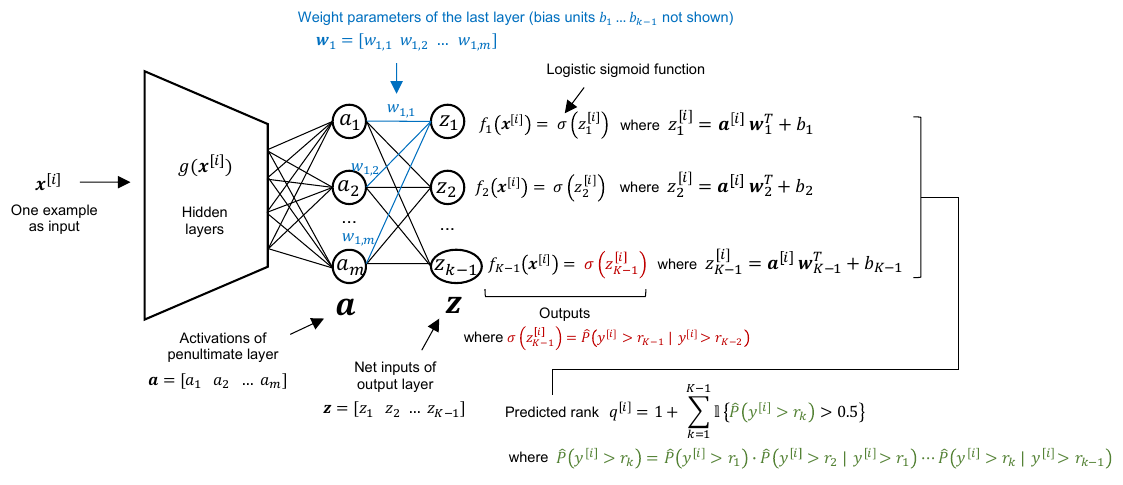}}
\caption{Outline of the neural network architecture used for CORN.}
\label{fig:architecture}
\end{center}
\end{figure*}

The proposed CORN model is a neural network for ordinal regression that {guarantees} rank consistency without any weight-sharing constraint in the output layer (Fig.~\ref{fig:architecture}). Instead, CORN uses a new training procedure with conditional training subsets that ensures rank consistency through applying the chain rule of probability.

\subsection{Rank-consistent Ordinal Regression based on Conditional Probabilities}
\label{sec:conditional-probabilities}

Given a training set ${D=\left\{\mathbf{x}^{[i]},y^{[i]}\right\}_{i=1}^N}$, CORN applies a label extension to the rank labels $y^{[i]}$ similar to CORAL, such that the resulting binary label ${y^{[i]}_{k} \in \{0,1\}}$ indicates whether $y^{[i]}$ exceeds rank $r_k$. Similar to CORAL, CORN also uses $K-1$ learning tasks associated with ranks $r_1, r_2, ..., r_K$ in the output layer as illustrated in Fig.~\ref{fig:architecture}. 

However, in contrast to CORAL, CORN estimates a series of conditional probabilities using conditional training subsets (described in Section~\ref{sec:conditional-training-set}) such that the output of the $k-$th binary task $f_k\left( \mathbf{x}^{[i]} \right)$ represents the conditional probability\footnote{When $k=1$, $f_k\left( \mathbf{x}^{[i]} \right)$  represents the initial unconditional probability $f_1\left( \mathbf{x}^{[i]} \right) = \hat{P}\left(y^{[i]} > r_1\right)$.}

\begin{equation}
\label{eq:cond}
    f_k\left( \mathbf{x}^{[i]} \right) = \hat{P}\left(y^{[i]} > r_k \,|\, y^{[i]} > r_{k-1}\right),
\end{equation}

\noindent where the events are nested: $\left\{y^{[i]} >r_{k}\right\} \subseteq \left\{y^{[i]} >r_{k-1}\right\}$.


The transformed, unconditional probabilities can then be computed by applying the chain rule for probabilities to the model outputs:

\begin{equation}
\label{eq:conditional-probabilities}
\hat{P}\left(y^{[i]} > r_k\right) = \prod^k_{j=1} f_j\left( \mathbf{x}^{[i]} \right).    
\end{equation}

\noindent Since  $\forall j, \; 0\leq f_j\left( \mathbf{x}^{[i]} \right) \leq 1$, we have 

\begin{equation}
\label{eq:rank-consistency}
    \hat{P}\left(y^{[i]} > r_1\right) \geq \hat{P}\left(y^{[i]} > r_2\right) \geq ... \geq \hat{P}\left(y^{[i]} > r_{K-1}\right),
\end{equation}

\noindent which guarantees rank consistency among the $K-1$ binary tasks.

\subsection{Conditional Training Subsets}
\label{sec:conditional-training-set}

Our model aims to estimate $f_1\left(\mathbf{x}^{[i]}\right)$ and the conditional probabilities $f_2\left(\mathbf{x}^{[i]}\right),...,f_{K-1}\left( \mathbf{x}^{[i]} \right)$. Estimating $f_1\left(\mathbf{x}^{[i]}\right)$ is a classic binary classification task under the extended binary classification framework with the binary labels $y_1^{[i]}$. To estimate the conditional probabilities such as $\hat{P}\left(y^{[i]} > r_2 \,|\, y^{[i]} > r_1\right)$, we focus only on the subset of the training data where $y^{[i]} > r_1$. As a result, when we minimize the binary cross-entropy loss on these conditional subsets, for each binary task, the estimated output probability has a proper conditional probability interpretation\footnote{When training a neural network using backpropagation, instead of minimizing the $K-1$ loss functions corresponding to the $K-1$ conditional probabilities on each conditional subset separately, we can minimize their sum, as shown in the loss function we propose in Section~\ref{sec:loss-function}, to optimize the binary tasks simultaneously.}.

In order to model the conditional probabilities in Eq.~\ref{eq:conditional-probabilities}, we construct conditional training subsets for training, which are used in the loss function (Section~\ref{sec:loss-function}) that is minimized via backpropagation. The conditional training subsets are obtained from the original training set as follows:

\begin{align*}
& S_1: \text{ all } \left\{ \left(\mathbf{x}^{[i]}, y^{[i]} \right) \right\}, \text{ for } i \in \{1, ..., N\},\\
& S_2: \left\{(\mathbf{x}^{[i]}, y^{[i]}) \; | \; y^{[i]} > r_1 \right\},\\
&\dots \\
& S_{K-1}: \left\{(\mathbf{x}^{[i]}, y^{[i]})\; | \; y^{[i]} > r_{k-2}\right\},
\end{align*}

\noindent where $N = |S_1| \geq |S_2| \geq ... \geq |S_{K-1}|$, and $|S_k|$ denotes the size of $S_k$. Note that the labels $y^{[i]}$ are subject to the binary label extension as described in Section~\ref{sec:conditional-probabilities}. Each conditional training subset $S_k$ is used for training the conditional probability prediction $\hat{P}\left(y^{[i]}>r_k \,|\, y^{[i]}>r_{k-1}\right)$ for $k\geq 2$.

{Additional theoretical justification for constructing the conditional training subsets is provided in the Supplementary Material in section \textit{Theoretical Analysis of Conditional Probability Estimation}. Section~\ref{sec:ablationstudy} compares the predictive performance of the CORN method with and without training subsets.}

\subsection{Loss Function}
\label{sec:loss-function}

Let $f_j\left(\mathbf{x}^{[i]}\right)$ denote the predicted value of the $j$-th node in the output layer of the network (Fig.~\ref{fig:architecture}), and let $|S_j|$ denote the size of the $j$-th conditional training set. To train a CORN neural network using backpropagation, we minimize the following loss function:

\begin{multline}
\label{eq:loss1}
L(\mathbf{X}, \mathbf{y}) = \\
 - \frac{1}{\sum_{j=1}^{K-1}|S_j|} \sum_{j=1}^{K-1}  \sum_{i=1}^{|S_j|} \bigg[\log\left(f_j(\mathbf{x}^{[i]})\right) \cdot \mathbbm{1}\left\{y^{[i]} > r_j\right\} \\ + \log\left(1-f_j\left(\mathbf{x}^{[i]}\right)\right) \cdot \mathbbm{1}\left\{y^{[i]} \leq r_j\right\} \bigg],
\end{multline}
We note that in $f_j(\mathbf{x}^{[i]})$, $\mathbf{x}^{[i]}$ represents the $i$-th training example in $S_j$. To simplify the notation, we omit an additional index $j$ to distinguish between $\mathbf{x}^{[i]}$ in different conditional training sets.

To improve the numerical stability of the loss gradients during training, we implement the following alternative formulation of the loss, where $\mathbf{Z}$ are the net inputs of the last layer (aka logits), as shown in Fig.~\ref{fig:architecture}, and $ \log\left(\sigma\left(\mathbf{z}^{[i]}\right)\right) = \log\left(f_j\left(\mathbf{x}^{[i]}\right)\right)$:

\begin{multline}
L(\mathbf{Z}, \mathbf{y}) = \\
- \frac{1}{\sum_{j=1}^{K-1}|S_j|} \sum_{j=1}^{K-1}  \sum_{i=1}^{|S_j|} \bigg[\log\left(\sigma\left(\mathbf{z}^{[i]}\right)\right) \cdot \mathbbm{1}\left\{y^{[i]} > r_j\right\}
\\+
\left(\log\left(\sigma\left(\mathbf{z}^{[i]}\right)\right)- \mathbf{z}^{[i]}\right) \cdot \mathbbm{1}\left\{y^{[i]} \leq r_j\right\} \bigg].
\end{multline}

\noindent A derivation showing that the two loss equations are equivalent and a PyTorch implementation are included in the Supplementary Material {in the section \textit{Numerically Stable Loss Function}}. In addition, the Supplementary Material includes a visual illustration of the loss computation based on the conditional training subsets ({Figure \ref{fig:corn-loss}}) and a {theoretical \textit{Generalization Bounds} analysis}.

\subsection{Rank Prediction}

To obtain the rank index $q$ of the $i$-th training example, and any new data record during inference, we threshold the predicted probabilities corresponding to the $K-1$ binary tasks and sum the binary labels as follows:

$$
q^{[i]}=1+\sum_{j=1}^{K-1} \mathbbm{1} \left(\hat{P}\left(y^{[i]}>r_j\right) > 0.5 \right),
$$
\noindent where the predicted rank is $r_{q^{[i]}}$.

\section{Experiments}
\label{sec:experiments}

\subsection{Datasets and Preprocessing}
\label{sec:datasets}

 The {MORPH-2} dataset\footnote{\url{https://www.faceaginggroup.com/morph/}}~\cite{ricanek2006morph} contains 55,608 face images, which were processed as described in~\cite{cao2020rank}: facial landmark detection~\cite{sagonas2016300} was used to compute the average eye location, which was then used by the EyepadAlign function in MLxtend v0.14~\cite{raschka2018mlxtend} to align the face images. The original {MORPH-2} dataset contains age labels in the range of 16-70 years. In this study, we use a balanced version of the {MORPH-2} dataset containing 20,625 face images with 33 evenly distributed age labels within the range of 16-48 years.

The Asian Face dataset (AFAD)\footnote{\url{https://github.com/afad-dataset/tarball}}~\cite{niu2016ordinal} contains 165,501 faces in the age range of 15-40 years. No additional preprocessing was applied to this dataset since the faces were already centered. In this study, we use a balanced version of the AFAD dataest with 13 age labels in the age range of 18-30 years. 

The Image Aesthetic dataset (AES)\footnote{\url{http://www.di.unito.it/~schifane/dataset/beauty-icwsm15/}}~\cite{schifanella2015image} used in this study contains 13,868 images, each with a list of beauty scores ranging from 1 to 5. To create ordinal regression labels, we replaced the beauty score list of each image with its average score rounded to the nearest integer in the range 1-5. Compared to the other image datasets {MORPH-2} and AFAD, the size of the AES dataset was relatively small, and we did not attempt to create a class-balanced version of this dataset for this study.

The Fireman dataset (Fireman)\footnote{\url{https://github.com/gagolews/ordinal_regression_data}} is a tabular dataset that contains 40,768 instances, 10 numeric features, and an ordinal response variable with 16 categories. We created a balanced version of this dataset consisting of 2,543 instances per class and 40,688 from the 16 ordinal classes in total.

Each dataset was randomly divided into 75\% training data, 5\% validation data, and 20\% test data. We share the partitions for all datasets, along with all preprocessing code used in this paper, in the code repository (see Section~\ref{sec:hardware-and-software}).

\subsection{Neural Network Architectures}
\label{sec:architecture}

\subsubsection{{Comparison with binary label extension frameworks for ordinal regression}}

For {the main method} comparisons {to other binary extension frameworks for ordinal regression} on the image datasets ({MORPH-2} and AFAD,), we used ResNet34~\cite{he2016deep} as the backbone architecture since it is an established architecture that is known to achieve good performance on a variety of image classification datasets. Besides the hyperparameter settings listed in Tables~\ref{tab:hparam-table} and \ref{tab:best-hyperparam}; we adopt all other settings from the ResNet34 paper.

For the tabular Fireman dataset, we used a simple multilayer perceptron architecture (MLP) with leaky ReLU~\cite{maas2013rectifier} activation functions (negative slope 0.01). Since the MLP architectures were prone to overfitting, a dropout layer with drop probability 0.2 was added after the leaky ReLU activations in each hidden layer. In addition, we used the AdamW~\cite{loshchilov2017decoupled} optimizer with a weight decay rate of 0.2. The number of hidden layers (one or two) and the number of units per hidden layer were determined by hyperparameter tuning (see Section~\ref{sec:training} for more details). 

In this paper, we {focus on} comparing the performance of a neural network trained via the rank-consistent CORN approach to {the two prominent binary extension-based ordinal regression frameworks for deep learning}, the Niu et al.~\cite{niu2016ordinal} {OR-NN} method (no rank consistency) and CORAL (rank consistency by using identical weight parameters for all nodes in the output layer). As a performance baseline, we implement neural network classifiers trained with standard multicategory cross-entropy loss as a baseline, which we refer to as {CE-NN}. While all methods (CE-NN, OR-NN, CORAL, and CORN) use different loss functions during training, it is worth emphasizing that they can share similar backbone architectures and only require small changes in the output layer. For instance, to implement a neural network for ordinal regression using the proposed CORN method, we replaced the network's output layer with the corresponding binary conditional probability task layer.

\subsection{Training and Evaluation}
\label{sec:training}

{The} model evaluations and comparisons are based on the mean absolute error (MAE) and root mean squared error (RMSE), which are defined as follows:

\begin{align*}
\mathrm{MAE}=\frac{1}{N}\sum_{i=1}^{N}\left|y_{i}-h\left(\mathbf{x}_{i}\right)\right| \quad \text{ and }
\\ \mathrm{RMSE}=\sqrt{\frac{1}{N}\sum_{i=1}^{N}\left(y_{i}-h\left(\mathbf{x}_{i}\right)\right)^{2}},
\end{align*}
where $y^{[i]}$ is the ground truth rank of the $i$-th test example and $h(\mathbf{x}^{[i]})$ is the predicted rank, respectively.

 Then, using the best hyperparameter setting for each method, we repeated the model training five times using different random seeds (0, 1, 2, 3, and 4) for the random weight initialization and dataset shuffling. {We considered the exact same hyperparameter ranges for each method. (A detailed list of the hyperparameter configurations we considered is shown in Table~\ref{tab:hparam-table}.) Then, we selected the best hyperparameter configuration, using grid search, based on its validation set performance for each method before computing the test set performance. Note that both the hyperparameter configuration and the best training epoch were determined based on the validation set before computing the final model performance on the independent test set. The best hyperparameter values for each method are listed in Table~\ref{tab:best-hyperparam}.}

\begin{table*}[htb]
  \renewcommand*{\arraystretch}{1.4}
\begin{center}
	\caption{{Configurations for hyperparameter tuning.}}
\label{tab:hparam-table}
\footnotesize
\begin{tabular}{|l|c|c|c|c|} 
\hline
Backbone  & Learning rates & Batch sizes & Layer sizes\\ 
\hline
ResNet34 &
 5e{-5}, 1e{-4}, 2.5e{-4}, 5e{-4}, 1e{-3}, 5e{-3} & 16, 32, 64, 128, 256, 512 & NA\\
 \hline
MLP &
 1e{-5}, 5e{-5}, 1e{-4}, 5e{-4}, 1e{-3}, 5e{-3} & 16, 32, 64, 128, 256, 512  & Layer1: 50, 100, 200, 300  \\
 & & & Layer2: 50, 100, 200, 300\\

\hline
\end{tabular}
\label{tab:hyperparameters}

\end{center}
\end{table*}

\begin{table*}[htb]
  \renewcommand*{\arraystretch}{1.4}
\begin{center}
	\caption{{Best hyperparameter settings for image and tabular datasets. }}
\label{tab:best-hyperparam}
\scalebox{0.85}{
\begin{tabular}{|l|c|c|c|c|c|c|c|} 
\hline
Datasets & Backbone & Methods  & Learning rates & Batch sizes  & Number of layers & Layer hidden units \\ 
\hline
Image Datasets & ResNet34 & CE-NN &
 5e{-4} & 256  & - & -\\
 \hline
Image Datasets & ResNet34  & OR-NN &
 5e{-4} & 256  & - & - \\
 \hline
Image Datasets & ResNet34 & CORAL &
 5e{-4} & 256  & - & -\\
 \hline
Image Datasets & ResNet34 & CORN &
 5e{-4} & 16  & - & -\\
\hline
Fireman & MLP & CE-NN & 5e{-4} & 64  & 2 & $300\times 200$ \\
\hline
Fireman & MLP & OR-NN & 5e{-4} & 128  & 2 & $300\times 300$ \\
\hline
Fireman & MLP & CORAL & 5e{-4} & 64  & 2 & $300\times 200$ \\
\hline
Fireman & MLP & CORN & 1e{-3} & 128 & 2 & $300\times 300$ \\
\hline
\end{tabular}}

\end{center}
\end{table*}

The models were trained {for 200 epochs} using stochastic gradient descent via adaptive moment estimation~\cite{kingma2015adam} with the default decay rates and {carefully checked for convergence such that training and validation MAE started to diverge and the validation MAE started to stagnate or decline}. The complete training logs for all methods are provided in the {code repository (Section~\ref{sec:hardware-and-software}).}


\subsection{Hardware and Software}
\label{sec:hardware-and-software}
All neural networks were implemented in PyTorch 1.8~\cite{paszke2019pytorch}. The models were trained on NVIDIA GeForce RTX 2080Ti graphics cards {on a private workstation as well as T4 graphics cards using the Grid.ai platform}. We make all source code used for the experiments available\footnote{\url{https://github.com/Raschka-research-group/corn-ordinal-neuralnet}} and provide a user-friendly implementation of CORN in the \texttt{coral-pytorch} Python package\footnote{\url{https://github.com/Raschka-research-group/coral-pytorch}}.


\section{Results and Discussion}
\label{sec:results}

To compare deep neural networks trained with our proposed CORN method to CORAL~\cite{cao2020rank}, Niu et al.'s OR-NN~\cite{niu2016ordinal}, and the baseline cross-entropy loss (CE-NN), we conducted a series of experiments on three image datasets and one tabular dataset. As detailed in Section~\ref{sec:architecture}, the experiments on the MORPH and AFAD image datasets were based on the {ResNet34} architecture. We used a multilayer perceptron for the tabular Fireman dataset.

An additional study using a VGG16 backbone pre-trained on ImageNet and comparisons with SORD and CNNPOR can be found in the Supplementary Material in section \textit{Comparison with Other Deep Learning Methods for Ordinal Regression}. In addition, results on text datasets and recurrent neural networks are included in the Supplementary Material in section  \textit{Additional Results on Text Datasets using Recurrent Neural Networks}.

As the {main} results in Table~\ref{tab:all-results} show, CORN outperforms all other binary label extension methods for ordinal regression on the {{MORPH-2} and AFAD image datasets and is tied with OR-NN on the Fireman tabular} dataset. We repeated the experiments {with} different random seeds for model weight initialization and data shuffling, which ensures that the results are not coincidental.

\begin{table*} [htb]
  \renewcommand*{\arraystretch}{1.5}
\begin{center}
	\caption{{Prediction errors on the test sets (lower is better). Each cell represents the average (AVG) and standard deviation (SD) for 5 random seeds runs. Best results are highlighted in bold. A ResNet34 backbone was used for the {MORPH-2} and AFAD image datasets. A multilayer perceptron backbone was used for the Fireman tabular dataset. The class labels in all datasets were balanced. The full table of all random seeds runs can be found in the Supplementary Materials (Table~\ref{tab:detailed-results-with-seeds}).}}
\label{tab:all-results}
\scalebox{0.80}{
\begin{tabular}{|l|c|c|c|c|c|c|c|c|} 
\hline
\multirow{1}{*}{Method} & Metrics format & \multicolumn{2}{c|}{{MORPH-2} (Balanced)} & \multicolumn{2}{c|}{AFAD (Balanced)} & 
\multicolumn{2}{c|}{Fireman}\\ 
\cline{3-8}
 & \multicolumn{1}{c|}{} & \multicolumn{1}{c|}{MAE} & \multicolumn{1}{c|}{RMSE} & MAE & \multicolumn{1}{c|}{RMSE} 
 & MAE & \multicolumn{1}{c|}{RMSE}\\
\hline

\multirow{1}{*}{\begin{tabular}[c]{@{}l@{}}CE-NN\end{tabular}} 
 & \multicolumn{1}{l|}{{ AVG$\pm$SD}} & \multicolumn{1}{l|}{3.73 $\pm$ 0.12} & \multicolumn{1}{l|}{5.04 $\pm$ 0.20 } & \multicolumn{1}{l|}{3.28 $\pm$ 0.04} & \multicolumn{1}{l|}{4.19 $\pm$ 0.06} & 
 \multicolumn{1}{l|}{0.80 $\pm$ 0.01}&
  \multicolumn{1}{l|}{1.14 $\pm$ 0.01}\\
\hline
\multirow{1}{*}{\begin{tabular}[c]{@{}l@{}}OR-NN {\small\cite{niu2016ordinal}} \end{tabular}}

& \multicolumn{1}{l|}{{ AVG$\pm$SD}} & \multicolumn{1}{c|}{3.13  $\pm$ 0.09} & \multicolumn{1}{c|}{4.23 $\pm$ 0.10} & \multicolumn{1}{c|}{2.85 $\pm$ 0.03} & \multicolumn{1}{c|}{3.48 $\pm$ 0.04} &
 \multicolumn{1}{c|}{\textbf{0.76 $\pm$ 0.01}}& \multicolumn{1}{c|}{\textbf{1.08 $\pm$ 0.01}}\\ 
\hline
\multirow{1}{*}{\begin{tabular}[c]{@{}l@{}}CORAL {\small\cite{cao2020rank}} \end{tabular}}

 & \multicolumn{1}{l|}{{ AVG$\pm$SD}} & \multicolumn{1}{l|}{2.99 $\pm$ 0.04 } & \multicolumn{1}{l|}{4.01 $\pm$ 0.03 } & \multicolumn{1}{l|}{2.99 $\pm$ 0.03} & \multicolumn{1}{l|}{3.70 $\pm$ 0.07 } &
  \multicolumn{1}{l|}{0.82 $\pm$ 0.01 } & 
 \multicolumn{1}{l|}{1.15 $\pm$ 0.01 }\\
\hline
\multirow{1}{*}{\begin{tabular}[c]{@{}l@{}}CORN (ours) \\ \end{tabular}}

 & \multicolumn{1}{l|}{{ AVG$\pm$SD}} & \multicolumn{1}{l|}{\textbf{2.98 $\pm$ 0.02} } & \multicolumn{1}{l|}{\textbf{3.99 $\pm$ 0.05} } & \multicolumn{1}{l|}{\textbf{2.81 $\pm$ 0.02}} & \multicolumn{1}{l|}{\textbf{3.46 $\pm$ 0.02} } &
 \multicolumn{1}{l|}{\textbf{0.76 $\pm$ 0.01} }& 
 \multicolumn{1}{l|}{\textbf{1.08 $\pm$ 0.01} }\\
\hline
\end{tabular}}
\label{tab:avg-std-results}
\end{center}
\end{table*}

It is worth noting that even though CORAL's rank consistency was found to be beneficial for model performance~\cite{cao2020rank}, it performs noticeably worse than OR-NN on the balanced {MORPH-2} and AFAD datasets. This might likely be due to CORAL's weight-sharing constraint in the output layer, which could affect the expressiveness of the neural networks and thus limit the complexity of what it can learn. In contrast the CORN method, which is also rank-consistent, performs better than OR-NN on {MORPH-2} and AFAD.

We found that OR-NN and CORN have identical performances on the tabular Fireman dataset (Table~\ref{tab:all-results}), outperforming both the CE-NN and CORAL in both test MAE and test RMSE. Here, the performances are relatively close, and the 16-category prediction task is relatively easy for a fully connected neural network regardless of the loss function.

\subsection{{Ablation Study}}
\label{sec:ablationstudy}

{Given the superior performance of CORN across several datasets, we studied the importance of the training subsets. In this ablation study, created an alternative CORN method without training subsets subsets. Here, the conditional probability of the $k-$th binary task is computed as }

\begin{equation}
    f_k\left( \mathbf{x}^{[i]} \right) = \hat{P}\left(y^{[i]} > r_k \,\right),
\end{equation}

\noindent {which is a modified version of Eq.~\ref{eq:cond}. Note that this modification results in meaningless probability scores, however, the rank consistency via Eq.~\ref{eq:rank-consistency} is still guaranteed since the probability scores are still computed via Eq.~\ref{eq:conditional-probabilities}, and each score cannot be greater than 1.}

{We shall note that the modified CORN method without training subsets sees at least as many training examples as the regular CORN method. This is because each task now has access to the full training batch rather than a subset.}

{As the results in Table~\ref{tab:ablationresults} show, the subsets do not only play a crucial role for yielding meaningful and theoretically justified rank probability values in CORN but they also improve the predictive performance. Across all datasets, with the exception of {MORPH-2}, the neural network trained with the regular CORN method outperforms the alternative version without subsets.}

\begin{table}[htb]
\begin{center}
\footnotesize
\caption{{MAE prediction errors on the test sets for the ResNet34 backbone. The class labels in all datasets were balanced. Best results are highlighted in bold.}}
\label{tab:ablationresults}
\begin{tabular}{l|l|l|}
\cline{2-3}
                                         & CORN                                                        & \begin{tabular}[c]{@{}l@{}}CORN\\ w/o subsets\end{tabular}  \\ \hline
\multicolumn{1}{|l|}{{MORPH-2}} & 2.98 $\pm$ 0.02                                            & \begin{tabular}[c]{@{}l@{}}\textbf{2.93 $\pm$  0.04}\end{tabular} \\ \hline
\multicolumn{1}{|l|}{AFAD}    & \begin{tabular}[c]{@{}l@{}}\textbf{2.81$\pm$  0.02}\end{tabular}  & \begin{tabular}[c]{@{}l@{}}3.06 $\pm$  0.02\end{tabular} \\ \hline
\multicolumn{1}{|l|}{AES}                & \begin{tabular}[c]{@{}l@{}}\textbf{0.43 $\pm$  0.01}\end{tabular} & \begin{tabular}[c]{@{}l@{}}0.68 $\pm$  0.01\end{tabular} \\ \hline
\multicolumn{1}{|l|}{Fireman}            & \begin{tabular}[c]{@{}l@{}} \textbf{0.76 $\pm$  0.01}\end{tabular} & \begin{tabular}[c]{@{}l@{}}0.81 $\pm$  0.01\end{tabular} \\ \hline
\end{tabular}
\end{center}
\end{table}

\section{Conclusions}
\label{sec:conclusions}

In this paper, we developed the rank-consistent CORN framework for ordinal regression via conditional training datasets. We used CORN to train convolutional and fully connected neural architectures on ordinal response variables. Our experimental results showed that the CORN method improved the predictive performance compared to the rank-consistent reference framework CORAL. While our experiments focused on image and tabular datasets, the generality of our CORN method allows it to be readily applied to other types of datasets to solve ordinal regression problems with various neural network structures.

\section{Acknowledgements}

This research was supported by the Office of the Vice Chancellor for Research and Graduate Education at the University of Wisconsin-Madison with funding from the Wisconsin Alumni Research Foundation.

\bibliographystyle{abbrv}
\bibliography{refs}

\FloatBarrier
\clearpage


\beginsupplement

\section{Supplementary Material}

\subsection{Theoretical Analysis of Conditional Probability Estimation}
Suppose we are interested in estimating a series of conditional probabilities
\begin{align*}
&f_1(\mathbf{x})=P\left(y>r_1  \; \vert \;  \mathbf{x}\right),\\ 
&f_2(\mathbf{x})=P\left(y>r_2 \; \vert \;  y>r_1,\mathbf{x} \right), \\
&...,\\
&f_{K-1}(\mathbf{x})=P\left(y>r_{K-1} \; \vert \;  >r_{k-2}, \mathbf{x} \right),
\end{align*}
with the observed dataset ${D=\left\{\mathbf{x}^{[i]},y^{[i]}\right\}_{i=1}^N}$, where $f_k(\mathbf{x})$ is the functional form of the neural network model outputs that depend on the neural network model weights. The likelihood of the model weights can be written as
\begin{equation}
    L = \prod_{j=1}^{K-1}\prod_{j=1}^{|S_j|} \left[f_j\left(\mathbf{x}^{[j]}\right)^{\mathbbm{1}\left\{y^{[i]} > r_j\right\}} \cdot \left(1-f_j\left(\mathbf{x}^{[j]}\right)\right)^{\mathbbm{1}\left\{y^{[i]} \leq r_j\right\}}\right].
\end{equation}
Hence, minimizing the loss function (Eq. \ref{eq:loss1}) is equivalent to solving the maximum likelihood estimate of the functional form representations of the conditional probabilities. This is also the theoretical justification that we construct the conditional training sets in the data preparation for the CORN loss function. Without using the conditional set in the loss function, the estimated probabilities do not have a conditional probability maximum likelihood interpretation. After solving the maximum likelihood estimates of the conditional probabilities, it is natural to use the probability chain rule to find the unconditional probabilities of exceeding rank $r_k$ in Eq. \ref{eq:conditional-probabilities} given each input $\mathbf{x}$.

\subsection{Generalization Bounds}

Analogous to CORAL~\cite{cao2020rank} and based on established generalization bounds for binary classification, Theorem~\ref{th:gener-error} shows that the final rank prediction by CORN generalizes well when the binary classification tasks generalize well.

\begin{theorem}[reduction of generalization error]\label{th:gener-error}
Let $\mathcal{C}$ be the cost matrix for the ordinal label predictions, where $\mathcal{C}_{y,y}=0$ and $\mathcal{C}_{y,r_k}>0$ for $k\neq y$. $P$ is the underlying distribution of $(\mathbf{x},y)$, i.e., $(\mathbf{x},y)\sim P$. Furthermore, let $h(\mathbf{x})$ be the model output yielding the predicted rank $r_q$; that is, $h(\mathbf{x})=r_q$. Let $y^{(k)} = \mathbbm{1}\{y>r_{k}\}$, and $\hat{y}^{(k)} = \mathbbm{1}\{\hat{P}(y>r_k)>0.5\} = \mathbbm{1}\{f_1f_2\ldots f_{k}>0.5\}$ be the prediction of $y^{(k)}$.  Given the binary classification tasks $\{f_k\}_{k=1}^{K-1}$, which we obtain from minimizing the loss in Eq.~\ref{eq:loss1}, and the rank-monotonic $\hat{y}_{k}$, we have

\begin{equation}\label{eq:gen-bound1}
\resizebox{.66\hsize}{!}{$\underset{(\mathbf{x},y)\sim P}{\mathbb{E}}\mathcal{C}_{y,h(\mathbf{x})}
      \leq 
     \sum_{k=1}^{K-1}\big|\mathcal{C}_{y,r_k}-\mathcal{C}_{y,r_{k+1}}\big| \underset{(\mathbf{x},y)\sim P}{\mathbb{E}}\mathbbm{1}\{\hat{y}^{(k)}\neq y^{(k)}\}$}.
\end{equation}

\end{theorem}
\begin{proof}
For any $\mathbf{x}\in \mathcal{X}$, by Eq.~\ref{eq:rank-consistency} we have
\begin{equation*}
    \hat{y}^{(1)}\geq \hat{y}^{(2)} \geq \ldots \geq \hat{y}^{(K-1)}.
\end{equation*}

\noindent If $h(\mathbf{x})=y$, then $\mathcal{C}_{y,h(\mathbf{x})}=0$.\\
If $h(\mathbf{x})=r_q\prec y=r_s$, then $q<s$. We have 
\begin{equation*}
\hat{y}^{(1)}=\hat{y}^{(2)}=\ldots=\hat{y}^{(q-1)}=1	
\end{equation*}
\noindent and 
\begin{equation*}
\hat{y}^{(q)}=\hat{y}^{(q+1)}=\ldots=\hat{y}^{(K-1)}=0.
\end{equation*}
\noindent Also, 
\begin{equation*}
y^{(1)}=y^{(2)}=\ldots=y^{(s-1)}=1
\end{equation*}
and 
\begin{equation*}
y^{(s)}=y^{(s+1)}=\ldots=y^{(K-1)}=0.
\end{equation*}
Thus, $\mathbbm{1}\{\hat{y}^{(k)}\neq y^{(k)}\}=1$ if and only if $q\leq k\leq s-1$. Since $\mathcal{C}_{y,y}=0,$

\begin{align*}
    \mathcal{C}_{y,h(\mathbf{x})} & =  \sum_{k=q}^{s-1}(\mathcal{C}_{y,r_k}-\mathcal{C}_{y,r_{k+1}})\cdot \mathbbm{1}\{\hat{y}^{(k)}\neq y^{(k)}\} \\
    & \leq  \sum_{k=q}^{s-1}\big|\mathcal{C}_{y,r_k}-\mathcal{C}_{y,r_{k+1}}\big|\cdot \mathbbm{1}\{\hat{y}^{(k)}\neq y^{(k)}\} \\
    & \leq  \sum_{k=1}^{K-1}\big|\mathcal{C}_{y,r_k}-\mathcal{C}_{y,r_{k+1}}\big|\cdot \mathbbm{1}\{\hat{y}^{(k)}\neq y^{(k)}\}.
\end{align*}

Similarly, if $h(x)=r_q\succ y=r_s$, then $q>s$ and
\begin{align*}
    \mathcal{C}_{y,h(\mathbf{x})}  &=  \sum_{k=s}^{q-1}(\mathcal{C}_{y,r_{k+1}}-\mathcal{C}_{y,r_{k}})\cdot \mathbbm{1}\{\hat{y}^{(k)}\neq y^{(k)}\} \\
    & \leq \sum_{k=1}^{K-1}\big|\mathcal{C}_{y,r_{k+1}}-\mathcal{C}_{y,r_{k}}\big|\cdot \mathbbm{1}\{\hat{y}^{(k)}\neq y^{(k)}\}.
\end{align*}
In any case, we have 
\begin{equation}
\label{eq:generalization_bd}
    \mathcal{C}_{y,h(\mathbf{x})}\leq \sum_{k=1}^{K-1}\big|\mathcal{C}_{y,r_k}-\mathcal{C}_{y,r_{k+1}}\big|\cdot \mathbbm{1}\{\hat{y}^{(k)}\neq y^{(k)}\}.
\end{equation}

By taking the expectation on both sides with $(\mathbf{x},y)\sim P$, we arrive at Eq.~\ref{eq:gen-bound1}.
\end{proof}

\subsection{{Comparison with Other Deep Learning Methods for Ordinal Regression}}
\label{sec:other-methods}

{We compare CORN with two additional, recent ordinal regression methods that do not rely on the binary extension framework:
\begin{enumerate}
    \item the convolutional neural network with pairwise regularization for ordinal regression (CNNPOR) method by Liu, Long, and Goh~\cite{liu2018constrained};
    \item the \textit{soft ordinal vectors} (SORD) method by Diaz and Marathe~\cite{diaz2019soft}.
\end{enumerate}}
{To facilitate a fair comparison, we adopted the exact same architecture and preprocessing steps from~\cite{diaz2019soft} and~\cite{liu2018constrained}. Similar to CNNPOR and SORD, we used a VGG16~\cite{simonyan2014very} backbone pre-trained on ImageNet~\cite{deng2009imagenet} where only the last layer (output layer) was re-initialized with random weights following. Also, following the preprocessing steps in CNNPOR and SORD, the training images in the AES dataset were resized to $256\times256$ pixels and randomly cropped to $224\times224$ as well as randomly flipped across the horizontal axis.}

{As these additional results on the AES dataset show, CORN also outperforms other recent ordinal regression methods for deep learning (CNNPOR~\cite{liu2018constrained} and SORD~\cite{diaz2019soft}) overall when trained with a VGG16 backbone that was pre-trained on ImageNet (Table~\ref{tab:vgg16}).}

\begin{table*}[htb!]
\begin{center}
	\caption{{Best hyperparameter settings for the AES datasets. For CNNPOR~\cite{liu2018constrained} and SORD~\cite{diaz2019soft} settings, please refer to the respective papers.}}
\label{tab:best-hyperparam2}
\footnotesize
\begin{tabular}{|l|c|c|c|c|c|} 
\hline
Datasets & Backbone & Methods  & Learning rates & Batch sizes   \\ 
\hline
AES Nature, Animals, Urban, People & VGG16 & CE-NN &  5e{-5}, 5e{-5}, 5e{-5}, 5e{-5} & 32, 32, 32, 16 \\
\hline
AES Nature, Animals, Urban, People & VGG16 & OR-NN &  1e{-4}, 1e{-4}, 1e{-4}, 5e{-4} & 32, 32, 16, 16 \\
\hline
AES Nature, Animals, Urban, People & VGG16 & CORAL &  5e{-4}, 1e{-3}, 1e{-3}, 5e{-4} & 16, 16, 16, 32  \\
\hline
AES Nature, Animals, Urban, People & VGG16 & CORAL & 5e{-5}, 5e{-5}, 5e{-5}, 5e{-5} & 64, 64, 64, 32  \\
\hline

\end{tabular}

\end{center}
\end{table*}

\begin{table*}[htb!]
\begin{center}
	\caption{{Prediction errors on the test sets for the VGG16 backbone pre-trained on ImageNet. Best results are highlighted in bold.}}
\label{tab:vgg16}
\scalebox{0.7}{
\begin{tabular}{|l|llllll|}
\hline
\multirow{2}{*}{} & \multicolumn{6}{c|}{MAE  (lower is better)}                                                                                                                                                                                                                                                                                                                                                                                 \\ \cline{2-7} 
                  & \multicolumn{1}{l|}{CE-NN} & \multicolumn{1}{l|}{OR-NN~\cite{niu2016ordinal}}  & \multicolumn{1}{l|}{CORAL~\cite{cao2020rank}} & \multicolumn{1}{l|}{\begin{tabular}[c]{@{}l@{}}CORN\\ (ours)\end{tabular}} & \multicolumn{1}{l|}{\begin{tabular}[c]{@{}l@{}}CNNPOR~\cite{liu2018constrained}\end{tabular}} & \begin{tabular}[c]{@{}l@{}}SORD~\cite{diaz2019soft}\end{tabular} \\ \hline
Nature            & \multicolumn{1}{l|}{0.29}         & \multicolumn{1}{l|}{0.29}                                                               & \multicolumn{1}{l|}{0.28}      & \multicolumn{1}{l|}{0.29}                                                      & \multicolumn{1}{l|}{0.29}                                                               & \multicolumn{1}{l|}{\textbf{0.27}}                                                             \\ \hline
Animals            & \multicolumn{1}{l|}{0.28}          & \multicolumn{1}{l|}{\textbf{0.25}}                                                               & \multicolumn{1}{l|}{0.30}      & \multicolumn{1}{l|}{0.26}                                                      & \multicolumn{1}{l|}{0.32}                                                               & 0.31                                                             \\ \hline
Urban             & \multicolumn{1}{l|}{0.26}         & \multicolumn{1}{l|}{0.27}                                                               & \multicolumn{1}{l|}{0.27}      & \multicolumn{1}{l|}{\textbf{0.25}}                                                      & \multicolumn{1}{l|}{0.33}                                                               & 0.28                                                             \\ \hline
People            & \multicolumn{1}{l|}{0.29}       & \multicolumn{1}{l|}{0.28}                                                               & \multicolumn{1}{l|}{0.29}      & \multicolumn{1}{l|}{\textbf{0.26}}                                                      & \multicolumn{1}{l|}{0.32}                                                               & 0.31                                                             \\ \hline 
Overall           & \multicolumn{1}{l|}{0.28}      & \multicolumn{1}{l|}{\textbf{0.27}}                                                               & \multicolumn{1}{l|}{0.29}      & \multicolumn{1}{l|}{\textbf{0.27}}                                                      & \multicolumn{1}{l|}{0.32}                                                               & 0.29                                                             \\ \hline
\end{tabular}}
\end{center}
\end{table*}

\begin{table*}[htb]
\begin{center}
	\caption{Prediction errors on the test sets (lower is better). A ResNet34 backbone was used for the {MORPH-2} and AFAD image datasets. A multilayer perceptron backbone was used for the AES and Fireman tabular datasets. The class labels in all datasets were balanced. Best results are highlighted in bold.}
\label{tab:detailed-results-with-seeds}
\scalebox{0.8}{
\begin{tabular}{|l|c|c|c|c|c|c|c|c|}
\hline
\multirow{2}{*}{Method} & \multicolumn{1}{c|}{\multirow{2}{*}{\begin{tabular}[c]{@{}c@{}}Seed\end{tabular}}} & \multicolumn{2}{c|}{{MORPH-2}} & \multicolumn{2}{c|}{AFAD} & 
\multicolumn{2}{c|}{Fireman}\\ 
\cline{3-8}
 & \multicolumn{1}{c|}{} & \multicolumn{1}{c|}{MAE} & \multicolumn{1}{c|}{RMSE} & MAE & \multicolumn{1}{c|}{RMSE} 
 & MAE & \multicolumn{1}{c|}{RMSE} \\
\hline
\multirow{4}{*}{\begin{tabular}[c]{@{}l@{}}CE-NN\end{tabular}} 
 & 0 & 3.81 & 5.19 & 3.31 & 4.27  & 0.80 & 1.14\\
 & 1 & 3.60 & 4.8 & 3.28 & 4.19  & 0.80 & 1.14\\
 & 2 & 3.61 & 4.84 & 3.32 & 4.22  & 0.79 & 1.13\\ 
 & 3 & 3.85 & 5.21 & 3.24 & 4.15  & 0.80 & 1.16\\
 & 4 & 3.80 & 5.14 & 3.24 & 4.13  & 0.80 & 1.15\\
\cline{2-8}
 & \multicolumn{1}{l|}{{ AVG$\pm$SD}} & \multicolumn{1}{l|}{3.73 $\pm$ 0.12} & \multicolumn{1}{l|}{5.04 $\pm$ 0.20 } & \multicolumn{1}{l|}{3.28 $\pm$ 0.04} & \multicolumn{1}{l|}{4.19 $\pm$ 0.06} & 
 \multicolumn{1}{l|}{0.80 $\pm$ 0.01}&
  \multicolumn{1}{l|}{1.14 $\pm$ 0.01}\\
\hline
\multirow{4}{*}{\begin{tabular}[c]{@{}l@{}}OR-NN\\ {\small\cite{niu2016ordinal}} \end{tabular}} 
 & 0 & 3.21 & 4.25 & 2.81 & 3.45  & 0.75 & 1.07\\
 & 1 & 3.16 & 4.25 & 2.87 & 3.54  & 0.76 & 1.08\\
 & 2 & 3.16 & 4.31 & 2.82 & 3.46  & 0.77 & 1.10\\ 
  & 3 & 2.98 & 4.05 & 2.89 & 3.49 & 0.76 & 1.08\\
 & 4 & 3.13 & 4.27 & 2.86 & 3.45 & 0.74 & 1.07 \\
\cline{2-8}
& \multicolumn{1}{l|}{{ AVG$\pm$SD}} & \multicolumn{1}{c|}{3.13  $\pm$ 0.09} & \multicolumn{1}{c|}{4.23 $\pm$ 0.10} & \multicolumn{1}{c|}{2.85 $\pm$ 0.03} & \multicolumn{1}{c|}{3.48 $\pm$ 0.04} &
 \multicolumn{1}{c|}{\textbf{0.76 $\pm$ 0.01}}& \multicolumn{1}{c|}{\textbf{1.08 $\pm$ 0.01}}\\ 
\hline
\multirow{4}{*}{\begin{tabular}[c]{@{}l@{}}CORAL\\ {\small\cite{cao2020rank}} \end{tabular}} 
 & 0 & 2.94 & 3.98 & 2.95 & 3.60 & 0.82 & 1.14 \\
 & 1 & 2.97 & 4.03 & 2.99 & 3.69 &  0.83 & 1.16\\
 & 2 & 3.01 & 3.98 & 2.98 & 3.70 &  0.81 & 1.13\\ 
  & 3 & 2.98 & 4.01 & 3.00 & 3.78 &  0.82 & 1.16 \\
 & 4 & 3.03 & 4.06 & 3.04 & 3.75 &  0.82 & 1.15\\
\cline{2-8}
 & \multicolumn{1}{l|}{{ AVG$\pm$SD}} & \multicolumn{1}{l|}{2.99 $\pm$ 0.04 } & \multicolumn{1}{l|}{4.01 $\pm$ 0.03 } & \multicolumn{1}{l|}{2.99 $\pm$ 0.03} & \multicolumn{1}{l|}{3.70 $\pm$ 0.07 } &
  \multicolumn{1}{l|}{0.82 $\pm$ 0.01 } & 
 \multicolumn{1}{l|}{1.15 $\pm$ 0.01 }\\
\hline
\multirow{4}{*}{\begin{tabular}[c]{@{}l@{}}CORN\\ (ours)  \end{tabular}} 
 & 0 & 2.98 & 4 & 2.80 & 3.45  & 0.75 & 1.07\\
 & 1 & 2.99 & 4.01 & 2.81 & 3.44 & 0.76 & 1.08 \\
 & 2 & 2.97 & 3.97 & 2.84 & 3.48 &  0.77 & 1.10 \\ 
 & 3 & 3.00 & 4.06 & 2.80 & 3.48 &  0.76 & 1.08 \\
 & 4 & 2.95 & 3.92 & 2.79 & 3.45 &  0.74 & 1.07 \\
\cline{2-8}
 & \multicolumn{1}{l|}{{ AVG$\pm$SD}} & \multicolumn{1}{l|}{\textbf{2.98 $\pm$ 0.02} } & \multicolumn{1}{l|}{\textbf{3.99 $\pm$ 0.05} } & \multicolumn{1}{l|}{\textbf{2.81 $\pm$ 0.02}} & \multicolumn{1}{l|}{\textbf{3.46 $\pm$ 0.02} } &
 \multicolumn{1}{l|}{\textbf{0.76 $\pm$ 0.01} }& 
 \multicolumn{1}{l|}{\textbf{1.08 $\pm$ 0.01} }\\
\hline
\end{tabular}}
\end{center}
\end{table*}

\newpage 
\label{sec:rnn-results}
\subsection{Additional Results on Text Datasets using Recurrent Neural Networks}

This section describes additional results we obtained from comparing CORN to other methods on text datasets using recurrent neural networks (RNNs) with long short-term memory (LSTM) cells.

\begin{table*}[!htbp]
\begin{center}
	\caption{Prediction errors on the test sets for the RNN backbone (lower is better). The class labels for both the Coursera and TripAdvisor were balanced. Best results are highlighted in bold.}
\label{tab:rnn-results}
\footnotesize
\begin{tabular}{|l|c|c|c|c|c|c|c|c|c|}
\hline
\multirow{2}{*}{Method} & \multicolumn{1}{c|}{\multirow{2}{*}{\begin{tabular}[c]{@{}c@{}}Seed\end{tabular}}} & \multicolumn{2}{c|}{TripAdvisor} & \multicolumn{2}{c|}{Coursera} \\ 
\cline{3-6}
 & \multicolumn{1}{c|}{} & \multicolumn{1}{c|}{MAE} & \multicolumn{1}{c|}{RMSE} & MAE & \multicolumn{1}{c|}{RMSE} \\
\hline
\multirow{4}{*}{\begin{tabular}[c]{@{}l@{}}CE-RNN\end{tabular}} 
 & 0 & 1.13 & 1.56 & 1.01 & 1.48 \\
 & 1 & 1.04 & 1.53 & 0.97 & 1.05 \\
 & 2 & 1.05 & 1.54 & 1.12 & 1.65 \\ 
 & 3 & 1.23 & 1.81 & 1.18 & 1.76 \\
 & 4 & 1.03 & 1.52 & 0.84 & 1.26 \\
\cline{2-6}
 & \multicolumn{1}{l|}{{ AVG$\pm$SD}} & \multicolumn{1}{l|}{1.10 $\pm$ 0.09} & \multicolumn{1}{l|}{1.59 $\pm$ 0.12 } & \multicolumn{1}{l|}{1.02 $\pm$ 0.13} & \multicolumn{1}{l|}{1.53 $\pm$ 0.19} \\
\hline
\multirow{4}{*}{\begin{tabular}[c]{@{}l@{}}OR-RNN\\ {\small\cite{niu2016ordinal}} \end{tabular}} 
 & 0 & 1.06 & 1.53 & 0.98 & 1.34 \\
 & 1 & 1.09 & 1.50 & 0.93 & 1.24 \\
 & 2 & 1.11 & 1.53 & 1.12 & 1.47 \\ 
 & 3 & 1.23 & 1.52 & 1.11 & 1.53 \\
 & 4 & 1.07 & 1.40 & 0.85 & 1.16 \\
\cline{2-6}
 & \multicolumn{1}{l|}{{ AVG$\pm$SD}} & \multicolumn{1}{l|}{1.11 $\pm$ 0.07} & \multicolumn{1}{l|}{1.50 $\pm$ 0.06 } & \multicolumn{1}{l|}{1.00 $\pm$ 0.12} & \multicolumn{1}{l|}{1.35 $\pm$ 0.15} \\
\hline
\multirow{4}{*}{\begin{tabular}[c]{@{}l@{}}CORAL\\ {\small\cite{cao2020rank}} \end{tabular}} 
 & 0 & 1.15 & 1.58 & 0.99 & 1.29 \\
 & 1 & 1.14 & 1.49 & 1.03 & 1.39 \\
 & 2 & 1.16 & 1.46 & 1.14 & 1.40  \\ 
  & 3 & 1.19 & 1.41 & 1.20 & 1.40 \\
 & 4 & 1.13 & 1.47 & 0.82 & 1.11 \\
\cline{2-6}
 & \multicolumn{1}{l|}{{ AVG$\pm$SD}} & \multicolumn{1}{l|}{1.15 $\pm$ 0.02} & \multicolumn{1}{l|}{\textbf{1.48 $\pm$ 0.06 }} & \multicolumn{1}{l|}{1.04 $\pm$ 0.15} & \multicolumn{1}{l|}{\textbf{1.33 $\pm$ 0.13}} \\
\hline
\multirow{4}{*}{\begin{tabular}[c]{@{}l@{}}CORN\\ (ours)  \end{tabular}} 
 & 0 & 1.09 & 1.55 & 0.95 & 1.37 \\
 & 1 & 1.09 & 1.53 & 0.90 & 1.32 \\
 & 2 & 1.01 & 1.45 & 1.07 & 1.49 \\ 
 & 3 & 1.12 & 1.51 & 1.05 & 1.47  \\
 & 4 & 1.03 & 1.46 & 0.78 & 1.14 \\
\cline{2-6}
 & \multicolumn{1}{l|}{{ AVG$\pm$SD}} & \multicolumn{1}{l|}{\textbf{1.07 $\pm$ 0.05}} & \multicolumn{1}{l|}{1.50 $\pm$ 0.04 } & \multicolumn{1}{l|}{\textbf{0.95 $\pm$ 0.12}} & \multicolumn{1}{l|}{1.36 $\pm$ 0.14} \\
\hline
\end{tabular}
\end{center}
\end{table*}

Both the 100K Coursera's courses reviews dataset\footnote{\url{https://www.kaggle.com/septa97/100k-courseras-course-reviews-dataset}} and {TripAdvisor} hotels reviews dataset\footnote{\url{https://www.kaggle.com/andrewmvd/trip-advisor-hotel-reviews}} contain reviews with 5 rating labels ranging from 1 to 5 stars. We used balanced versions of these datasets to distribute the ratings evenly. The balanced Coursera dataset contains 11,852 reviews, and the TripAdvisor dataset contains 7,000 reviews. Each dataset was randomly divided into 75\% training data, 5\% validation data, and 20\% test data. The dataset splits and preprocessing code can be found in the code repository (see Section 4.4 of the main manuscript).

For method comparisons on the text datasets, we use a standard RNN with one LSTM cell. Similar to the image datasets, we compare the performance of a neural network trained via the rank-consistent CORN approach to both Niu et al.'s {OR-RNN} method (no rank consistency) and CORAL (rank consistency by using identical weight parameters for all nodes in the output layer). We also implemented RNN classifiers trained with standard multicategory cross-entropy loss as a baseline, which we refer to as {CE-RNN}. All methods share similar backbone architectures and only require minor changes in the output layer.

The training and evaluation steps are similar to those of the image datasets in the main manuscript. The RNN models were trained for 200 epochs using ADAM with default settings. The model with the best validation set performance was then chosen as the final model for evaluation on the test set. The training logs for all runs are available in the CORN GitHub repository (see Section 4.4). The learning rates considered for hyperparameter tuning were 1e{-5}, 5e{-5}, 1e{-4}, 5e{-4}, 1e{-3}, 5e{-3}, and we considered batch sizes 16, 32, 64, 128, 256, 512.

As the results in Table~\ref{tab:rnn-results} show, CORN outperforms all other methods on the two text datasets, TripAdvisor and Coursera, in terms of the test set MAE. All experiments were repeated for different random seeds to ensure that the results were not coincidental.

It is worth noticing that while the CORN method showed superior performance in terms of test MAE, the CORAL method performed better in test RMSE compared with all other methods. One possible explanation is that since RMSE penalizes large gaps more harshly than MAE, CORAL may behave slightly better on outliers while CORN may make fewer mistakes in total. However, both methods show reliable performance over the text datasets.

\newpage

\subsection{Detailed Performance Table}

Table\ref{tab:detailed-results-with-seeds} is a more detailed version of the results table shown in the main paper, listing the performance for each individual random seed.

\begin{table*}[htb!]
\begin{center}
	\caption{Prediction errors on the test sets (lower is better). A ResNet34 backbone was used for the {MORPH-2} and AFAD image datasets. A multilayer perceptron backbone was used for the AES and Fireman tabular datasets. The class labels in all datasets were balanced. Best results are highlighted in bold.}
\label{tab:detailed-results-with-seeds}
\scalebox{0.8}{
\begin{tabular}{|l|c|c|c|c|c|c|c|c|}
\hline
\multirow{2}{*}{Method} & \multicolumn{1}{c|}{\multirow{2}{*}{\begin{tabular}[c]{@{}c@{}}Seed\end{tabular}}} & \multicolumn{2}{c|}{{MORPH-2}} & \multicolumn{2}{c|}{AFAD} & 
\multicolumn{2}{c|}{Fireman}\\ 
\cline{3-8}
 & \multicolumn{1}{c|}{} & \multicolumn{1}{c|}{MAE} & \multicolumn{1}{c|}{RMSE} & MAE & \multicolumn{1}{c|}{RMSE} 
 & MAE & \multicolumn{1}{c|}{RMSE} \\
\hline
\multirow{4}{*}{\begin{tabular}[c]{@{}l@{}}CE-NN\end{tabular}} 
 & 0 & 3.81 & 5.19 & 3.31 & 4.27  & 0.80 & 1.14\\
 & 1 & 3.60 & 4.8 & 3.28 & 4.19  & 0.80 & 1.14\\
 & 2 & 3.61 & 4.84 & 3.32 & 4.22  & 0.79 & 1.13\\ 
 & 3 & 3.85 & 5.21 & 3.24 & 4.15  & 0.80 & 1.16\\
 & 4 & 3.80 & 5.14 & 3.24 & 4.13  & 0.80 & 1.15\\
\cline{2-8}
 & \multicolumn{1}{l|}{{ AVG$\pm$SD}} & \multicolumn{1}{l|}{3.73 $\pm$ 0.12} & \multicolumn{1}{l|}{5.04 $\pm$ 0.20 } & \multicolumn{1}{l|}{3.28 $\pm$ 0.04} & \multicolumn{1}{l|}{4.19 $\pm$ 0.06} & 
 \multicolumn{1}{l|}{0.80 $\pm$ 0.01}&
  \multicolumn{1}{l|}{1.14 $\pm$ 0.01}\\
\hline
\multirow{4}{*}{\begin{tabular}[c]{@{}l@{}}OR-NN\\ {\small\cite{niu2016ordinal}} \end{tabular}} 
 & 0 & 3.21 & 4.25 & 2.81 & 3.45  & 0.75 & 1.07\\
 & 1 & 3.16 & 4.25 & 2.87 & 3.54  & 0.76 & 1.08\\
 & 2 & 3.16 & 4.31 & 2.82 & 3.46  & 0.77 & 1.10\\ 
  & 3 & 2.98 & 4.05 & 2.89 & 3.49 & 0.76 & 1.08\\
 & 4 & 3.13 & 4.27 & 2.86 & 3.45 & 0.74 & 1.07 \\
\cline{2-8}
& \multicolumn{1}{l|}{{ AVG$\pm$SD}} & \multicolumn{1}{c|}{3.13  $\pm$ 0.09} & \multicolumn{1}{c|}{4.23 $\pm$ 0.10} & \multicolumn{1}{c|}{2.85 $\pm$ 0.03} & \multicolumn{1}{c|}{3.48 $\pm$ 0.04} &
 \multicolumn{1}{c|}{\textbf{0.76 $\pm$ 0.01}}& \multicolumn{1}{c|}{\textbf{1.08 $\pm$ 0.01}}\\ 
\hline
\multirow{4}{*}{\begin{tabular}[c]{@{}l@{}}CORAL\\ {\small\cite{cao2020rank}} \end{tabular}} 
 & 0 & 2.94 & 3.98 & 2.95 & 3.60 & 0.82 & 1.14 \\
 & 1 & 2.97 & 4.03 & 2.99 & 3.69 &  0.83 & 1.16\\
 & 2 & 3.01 & 3.98 & 2.98 & 3.70 &  0.81 & 1.13\\ 
  & 3 & 2.98 & 4.01 & 3.00 & 3.78 &  0.82 & 1.16 \\
 & 4 & 3.03 & 4.06 & 3.04 & 3.75 &  0.82 & 1.15\\
\cline{2-8}
 & \multicolumn{1}{l|}{{ AVG$\pm$SD}} & \multicolumn{1}{l|}{2.99 $\pm$ 0.04 } & \multicolumn{1}{l|}{4.01 $\pm$ 0.03 } & \multicolumn{1}{l|}{2.99 $\pm$ 0.03} & \multicolumn{1}{l|}{3.70 $\pm$ 0.07 } &
  \multicolumn{1}{l|}{0.82 $\pm$ 0.01 } & 
 \multicolumn{1}{l|}{1.15 $\pm$ 0.01 }\\
\hline
\multirow{4}{*}{\begin{tabular}[c]{@{}l@{}}CORN\\ (ours)  \end{tabular}} 
 & 0 & 2.98 & 4 & 2.80 & 3.45  & 0.75 & 1.07\\
 & 1 & 2.99 & 4.01 & 2.81 & 3.44 & 0.76 & 1.08 \\
 & 2 & 2.97 & 3.97 & 2.84 & 3.48 &  0.77 & 1.10 \\ 
 & 3 & 3.00 & 4.06 & 2.80 & 3.48 &  0.76 & 1.08 \\
 & 4 & 2.95 & 3.92 & 2.79 & 3.45 &  0.74 & 1.07 \\
\cline{2-8}
 & \multicolumn{1}{l|}{{ AVG$\pm$SD}} & \multicolumn{1}{l|}{\textbf{2.98 $\pm$ 0.02} } & \multicolumn{1}{l|}{\textbf{3.99 $\pm$ 0.05} } & \multicolumn{1}{l|}{\textbf{2.81 $\pm$ 0.02}} & \multicolumn{1}{l|}{\textbf{3.46 $\pm$ 0.02} } &
 \multicolumn{1}{l|}{\textbf{0.76 $\pm$ 0.01} }& 
 \multicolumn{1}{l|}{\textbf{1.08 $\pm$ 0.01} }\\
\hline
\end{tabular}}
\end{center}
\end{table*}

\newpage

\subsection{Numerically Stable Loss Function}

We can convert the CORN loss function,

\begin{multline}
L(\mathbf{X}, \mathbf{y}) = \\
 - \frac{1}{\sum_{j=1}^{k-1}|S_j|} \sum_{j=1}^{k-1}  \sum_{i=1}^{|S_j|} \bigg[\log\left(f_j(\mathbf{x}^{[i]})\right) \cdot \mathbbm{1}\left\{y^{[i]} > r_j\right\} \\ + \log\left(1-f_j\left(\mathbf{x}^{[i]}\right)\right) \cdot \mathbbm{1}\left\{y^{[i]} \leq r_j\right\} \bigg],
\end{multline}

\noindent into an alternative version

\begin{multline}
L(\mathbf{Z}, \mathbf{y}) = \\
- \frac{1}{\sum_{j}^{k-1}|S_j|} \sum_{j=1}^{k-1}  \sum_{i=1}^{|S_j|} \bigg[\log\left(\sigma\left(\mathbf{z}^{[i]}\right)\right) \cdot \mathbbm{1}\left\{y^{[i]} > r_j\right\}
\\+
\left(\log\left(\sigma\left(\mathbf{z}^{[i]}\right)\right)- \mathbf{z}^{[i]}\right) \cdot \mathbbm{1}\left\{y^{[i]} \leq r_j\right\} \bigg],
\end{multline}

\noindent where $\mathbf{Z}$ are the net inputs of the last layer (aka logits) and $ \log\left(\sigma\left(\mathbf{z}^{[i]}\right)\right) = \log\left(f_j\left(\mathbf{x}^{[i]}\right)\right)$, since 

\begin{align*}
&\log\left(1 - \frac{1}{1+e^{-z}} \right)\\
&= \log\left(1 - \frac{e^z}{1+e^z}\right)\\
&= \log\left(\frac{1}{1+e^z}\right)\\
&= \log\left(\frac{e^z}{1+e^z}\right) - \log(e^z)\\
&= \log\left( \frac{1}{1+e^{-z}} \right) - z\\
&= \log\left(\sigma(z) \right) -z.\\
\end{align*}

\begin{figure}[htb!]
\begin{center}
\centerline{\includegraphics[width=0.78\linewidth]{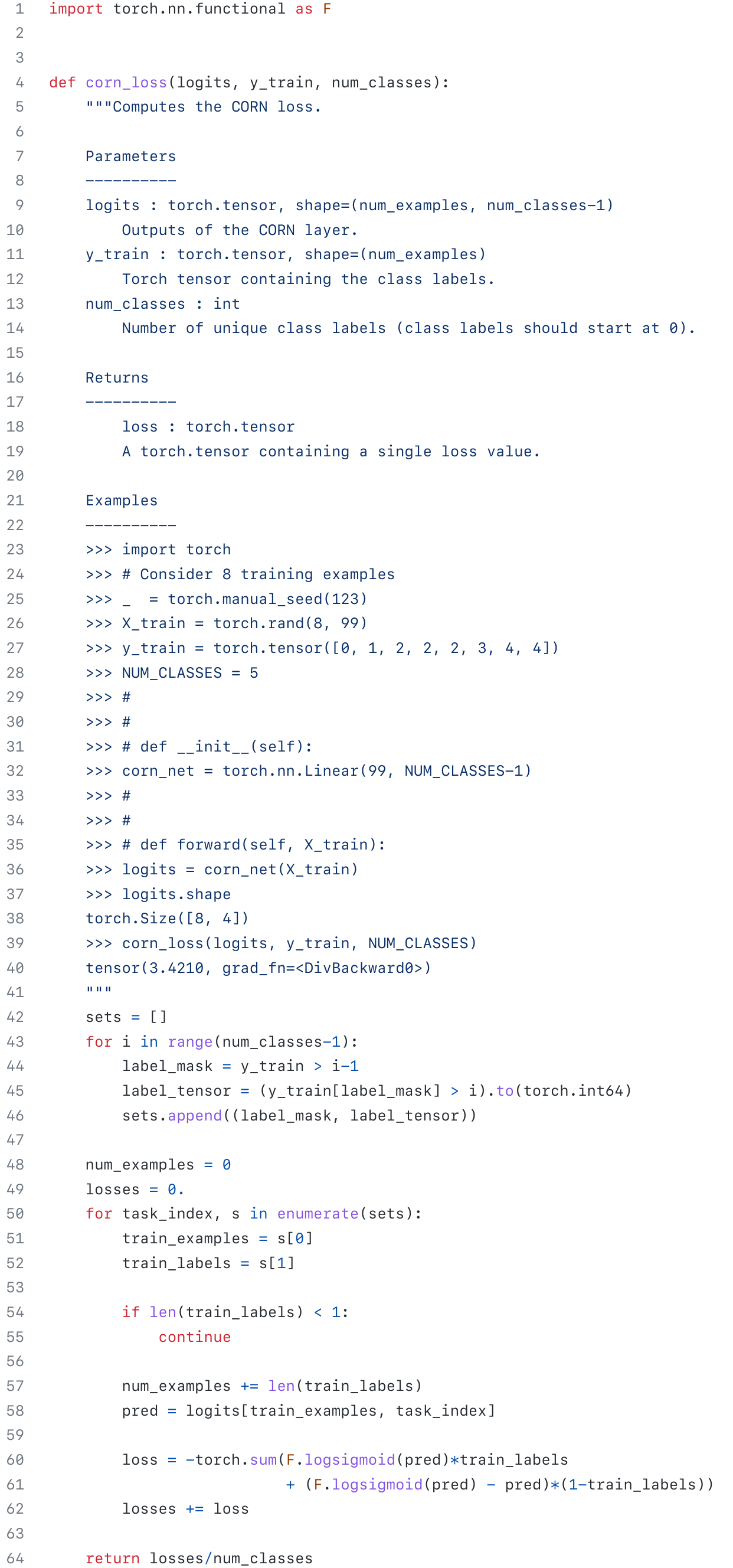}}
\caption{CORN loss function implemented in PyTorch v. 1.8.}
\label{fig:corn-loss}
\end{center}
\end{figure}

This allows us to use the \verb+logsigmoid(z)+ function that is implemented in deep learning libraries such as PyTorch as opposed to using \verb+log(1-sigmoid(z))+; the former yields numerically more stable gradients during backpropagation. A PyTorch implementation of the CORN loss function is shown in Fig.~\ref{fig:corn-loss}.

\FloatBarrier
\clearpage

\subsection{{Additional Figures Explaining the CORN Method}}

\begin{figure*}[htb!]
\begin{center}
\centerline{\includegraphics[width=0.98\linewidth]{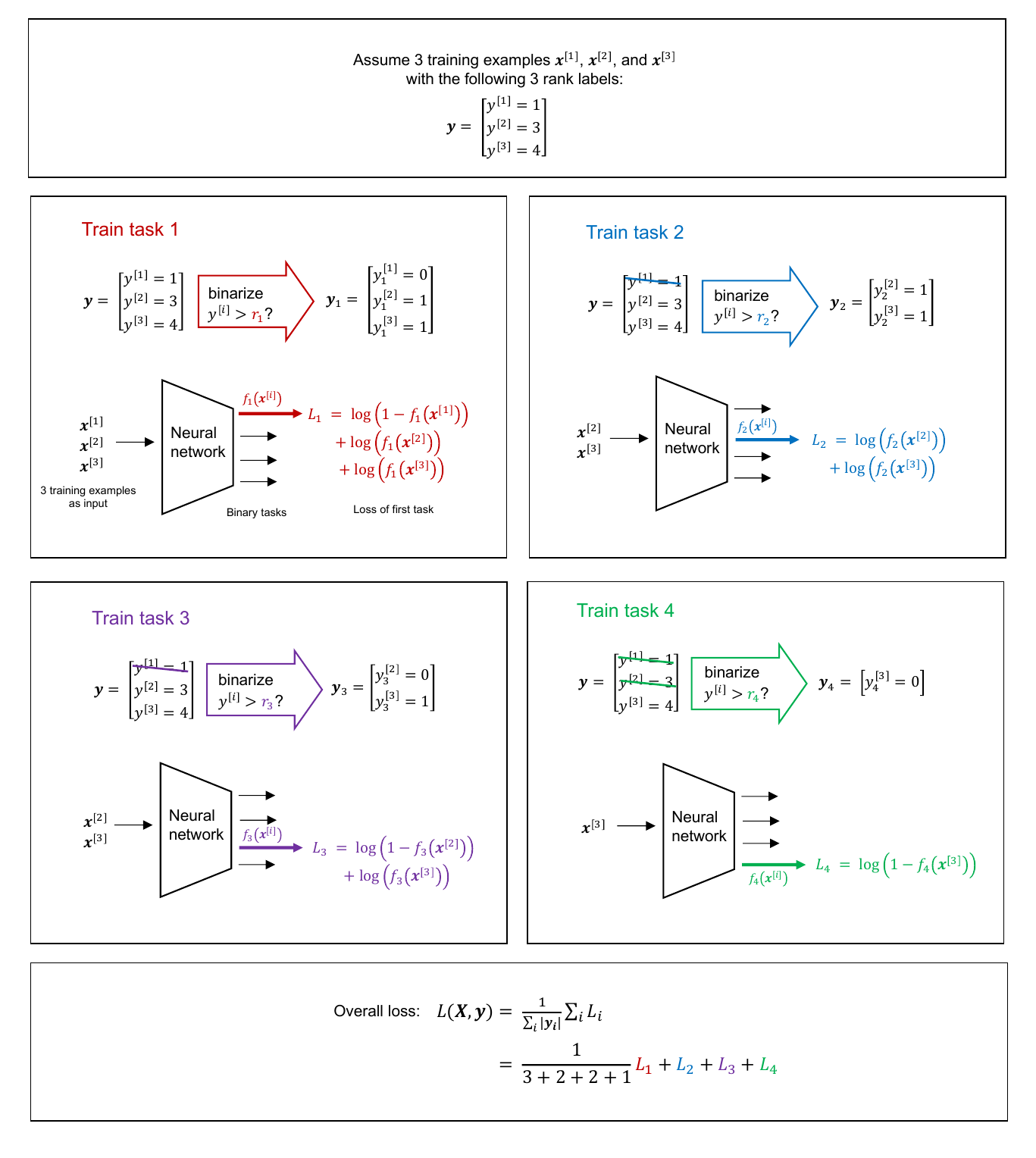}}
\caption{Visual explanation of how the CORN loss is computed using the conditional training subsets.}
\label{fig:training}
\end{center}
\end{figure*}

\begin{figure*}[htb!]
\begin{center}
\centerline{\includegraphics[width=0.98\linewidth]{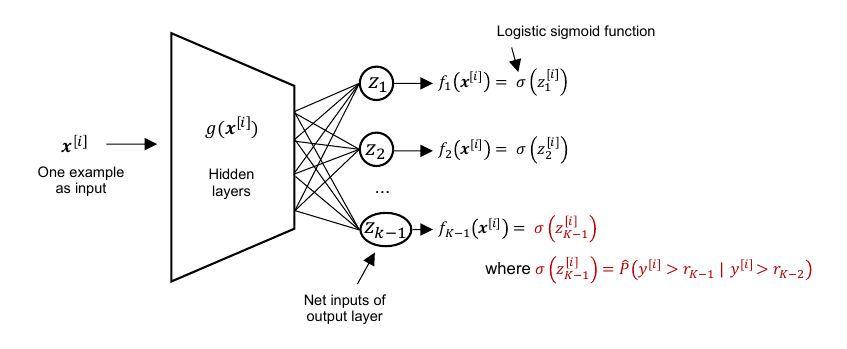}}
\caption{Outline of a neural network architecture that can be trained using CORN. Compared to a regular classification network, the only architecture modification is that the output layer consists of $k-1$ instead of $k$ nodes, where $k$ represents the number of unique ordinal labels in the dataset. The hidden layers represent the layers of an existing backbone architecture, such as a standard ResNet-34.}
\label{fig:architecture-simplified}
\end{center}
\end{figure*}

\end{document}